\documentclass[twoside]{article}

\usepackage[preprint]{aistats2026}
\usepackage{makecell}
\usepackage[utf8]{inputenc}
\usepackage[T1]{fontenc}
\usepackage{hyperref}
\usepackage{url}
\usepackage{booktabs}
\usepackage{amsfonts}
\usepackage{nicefrac}
\usepackage{microtype}
\usepackage{xcolor}
\usepackage{comment}
\usepackage{graphicx}
\usepackage{amsmath}
\usepackage{mathtools}
\usepackage{algorithm}
\usepackage{algpseudocode}
\usepackage{mathtools}
\usepackage{mathrsfs}
\usepackage{caption}
\usepackage{subcaption}
\usepackage{listings}
\usepackage{float}
\usepackage{multirow}
\begin{document}

% \runningtitle{I use this title instead because the last one was very long}
\runningtitle{Causality--$\Delta$}

% \runningauthor{Surname 1, Surname 2, Surname 3, ...., Surname n}
\runningauthor{Rezvan, Gille, Schauer, Torkar}

\twocolumn[

\aistatstitle{Causality--$\Delta$: Jacobian-Based Dependency Analysis in Flow Matching Models}

\aistatsauthor{Reza Rezvan$^1$ \And Gustav Gille$^1$ \And Moritz Schauer$^{1,2}$ \And Richard Torkar$^{1,2}$}

\aistatsaddress{$^1$ Chalmers University of Technology \\ $^2$ University of Gothenburg}]

% Follow https://x.com/TrevorABranch/status/620699527486373888

\begin{abstract}
Flow matching learns a velocity field that transports a base distribution to data. We study how small latent perturbations propagate through these flows and show that Jacobian-vector products (JVPs) provide a practical lens on dependency structure in the generated features. We derive closed-form expressions for the optimal drift and its Jacobian in Gaussian and mixture-of-Gaussian settings, revealing that even globally nonlinear flows admit local affine structure. In low-dimensional synthetic benchmarks, numerical JVPs recover the analytical Jacobians. In image domains, composing the flow with an attribute classifier yields an attribute-level JVP estimator that recovers empirical correlations on MNIST and CelebA. Conditioning on small classifier-Jacobian norms reduces correlations in a way consistent with a hypothesized common-cause structure, while we emphasize that this conditioning is not a formal $\mathrm{do}$-intervention. 
\end{abstract}

% \keywords{Flow Matching, Causality, Intervention, Jacobian, Perturbation Analysis, Generative Models}

\section{Introduction} 
\label{sec:introduction}
Diffusion and score-based models have become powerful generative frameworks, transforming simple distributions into complex data across domains~\cite{ho2020denoising, song2021scorebased}. Flow matching offers a deterministic alternative by learning a velocity field that transports a base distribution to data~\cite{lipman2023flow, albergo2023building}. Despite their empirical success, the internal dependency structure of these models remains difficult to interpret.

We study how small latent perturbations propagate through flow matching models and show that Jacobian-vector products (JVPs) provide a practical estimator of dependency structure in generated features. Our analysis connects conditional expectations, local linearization, and the geometry of learned flows. While this perspective is causally motivated, we explicitly distinguish conditioning from formal interventions in the sense of Pearl~\cite{pearl2009causality}.

Specifically, our contributions are:

\begin{itemize}
    \itemsep-0.1em 
    \item Derive closed-form drifts and Jacobians for Gaussian and mixture-of-Gaussian targets, revealing local affine structure in flow matching.
    \item Introduce a JVP-based estimator for dependency structure and validate it on synthetic benchmarks and MNIST pixel correlations.
    \item Apply the method to CelebA attributes and show that conditioning on small classifier-Jacobian norms reduces correlations in a way consistent with a common-cause hypothesis, while clarifying that this is not a $\mathrm{do}$-intervention.
\end{itemize}

\section{Related Work}
\label{sec:related_work}
Diffusion and score-based models have become a dominant approach to generative modeling, with continuous-time formulations connecting SDEs, probability flow ODEs, and score estimation \cite{ho2020denoising, song2021scorebased, song2020generative}. Flow matching offers a deterministic alternative that directly learns the velocity field transporting a base distribution to data, and is closely related to stochastic interpolants \cite{lipman2023flow, albergo2023building}. Our work builds on this line by studying how local perturbations propagate through learned flows and by deriving closed-form drift/Jacobian expressions in Gaussian and mixture settings.

From a causality perspective, representation learning has been proposed as a way to expose mechanistic structure in high-dimensional data \cite{scholkopf2021toward}. Complementary to this, Jacobian-based analyses have been used to encourage or measure disentanglement in generative models \cite{wei2021orthogonal, khrulkov2021disentangled}. We connect these threads by analyzing Jacobian-vector products of flow matching models and using them to estimate dependency structure in both synthetic and image domains, while explicitly distinguishing conditioning from formal interventions.

\section{Preliminaries}
\label{sec:prelim}
In this section, we introduce the flow matching framework and the fundamentals of causality. We will briefly introduce and explain these concepts then clearly state how we will use these frameworks.

\subsection{Flow Matching} \label{subsec:flow_matching}

Flow matching \cite{lipman2023flow} provides an elegant framework for learning continuous normalizing flows by directly modeling the velocity field that transports one distribution to another. Unlike diffusion models that rely on explicit noise schedules, flow matching learns a deterministic ODE that maps between distributions.

Given a source distribution $p_{\text{init}}$ (typically standard Gaussian noise) and a target distribution $p_{\text{data}}$, flow matching aims to learn a time-dependent vector field $\boldsymbol{v}_\theta(\mathbf{x}(t), t)$ that guides the transformation from samples of $p_{\text{init}}$ to samples of $p_{\text{data}}$.

Let $\boldsymbol\phi_t: \mathbb{R}^d \mapsto \mathbb{R}^d$ be a continuous family of maps with $\boldsymbol\phi_0 = \mathbf{I}_d$ and $\boldsymbol\phi_1$ pushing $p_{\text{init}}$ to $p_{\text{data}}$. The transport ODE is given by,

\begin{equation*}
\label{eq:flow_matching_ODE}
\frac{d}{dt}\boldsymbol\phi_t(\mathbf{x}) = \mathbf  v_t( \boldsymbol\phi_t(\mathbf{x})), \quad \boldsymbol\phi_0 = \mathbf{I}_d,
\end{equation*}

where $\boldsymbol v_t$ is a time-dependent vector field.

For linear interpolation between samples $\boldsymbol x_0 \sim p_{\text{init}}$ and $\boldsymbol x_1 \sim p_{\text{data}}$, defined as $\boldsymbol x_t = (1-t)\boldsymbol x_0 + t\boldsymbol x_1$, the optimal vector field can be shown to be~\cite{lipman2023flow},

\begin{equation}
\label{eq:optimal_flow}
\boldsymbol u(\mathbf{x_t}, t) = \mathbb{E}_{p(\boldsymbol x_0, \boldsymbol x_1|\mathbf{x}_t)}[\boldsymbol x_1 - \boldsymbol x_0],
\end{equation}

which is the conditional expectation of the difference between the paired source and target points (i.e., the trajectory from noise to data), given the interpolated point $\mathbf{x}_t$. The flow matching objective is then to learn a parameterized velocity field $\mathbf v_\theta$ that approximates this optimal field,

\begin{equation*}
\mathcal{L}(\theta) = \mathbb{E}_{t, \boldsymbol x_0, \boldsymbol x_1}\left[ \left\| \mathbf v_\theta(\mathbf{x}_t, t) - (\boldsymbol x_1 - \boldsymbol x_0) \right\|^2 \right].
\end{equation*}

Once trained, the model can be used to generate samples by integrating the learned ODE starting from noise samples.

\subsection{Fundamentals of Causality}
Causality extends beyond simple statistical association, providing a framework for understanding how variables influence each other. Although correlation identifies patterns of co-occurrence, causation establishes directional relationships where changes in one variable directly produce changes in another \cite{pearl2009causality}.

In the context of generative models, causality offers a lens through which we can interpret the relationship between generated features. When a generative process transforms noise into data, we can ask \emph{which latent dimensions causally influence generated features}.

\subsubsection{Pearl's Ladder of Causation}
Pearl's causality framework consists of three levels of increasing complexity:
\cite{pearl2018book}

\begin{enumerate}
    \itemsep0em
    \item
    \textbf{Association (Seeing)}: The ability to recognize patterns and correlations in observational data.
    $P(y|x)$ -- "What is the probability of observing feature $Y$ given feature $X$"
    \item \textbf{Intervention (Doing)}: The ability to predict the effect of deliberate actions or interventions.
   $P(y|\mathrm{do}(x))$ -- What is the probability of the observing feature $Y$ given a specific condition on $X$.
   \item \textbf{Counterfactuals (Imagining)}: The ability to reason about what would have happened under different circumstances.
   $P(y_x | x', y')$ -- "What would feature $Y$ have been if feature $X$ had been $x$, given that I observed $X=x'$ and $Y=y'$?"
\end{enumerate}

Our analysis focuses primarily on the first and second levels as we examine how (small) perturbations in latent space propagate through the generative model affecting the joint output distribution. Here, we are interested in using Pearl's ladder specifically on the output features, i.e., intervening and observing how output features change and affect each other. We will now introduce the formal definition of intervention that will be used.

\subsubsection{Formal Definition of Intervention}
\label{subsubsec:intervention}

Let $\mathcal{M} = (\mathbf{U}, \mathbf{V}, \mathcal{F}, P(\mathbf{U}))$ be a structural causal model (SCM) where:

\begin{itemize}
    \itemsep0em
    \item $\mathbf{U}$ is a set of latent (exogenous) variables with joint distribution $P(\mathbf{U})$.
    
    \item $\mathbf{V} = \{V_1, \dots, V_n\}$ is a set of observed (endogenous) variables.
    
    \item $\mathcal{F} = \{f_i\}$ is a collection of structural assignments $V_i = f_i\bigl(\mathrm{pa}(V_i), U_i\bigr)$, with $\mathrm{pa}(V_i)\subseteq \mathbf{V}\setminus\{V_i\}$.
\end{itemize}

The \emph{intervention} $\mathrm{do}(X = x')$ on a variable $X \in \mathbf V$ produces a modified model,

$$
\mathcal{M}_{\mathrm{do}(X = x')} = \left(\mathbf{U}, \mathbf{V}, \mathcal{F}_{\mathrm{do}(X = x')}, P(\mathbf{U})\right),
$$

where the original assignment for $X$ is replaced by the constant assignment $X := x'$. All other structural equations remain unchanged.

The \emph{interventional distribution} of any variable $Y\in\mathbf V$ is then defined as,

$$
P_{\mathcal{M}}\left(Y \mid \mathrm{do}(X = x')\right) := P_{\mathcal{M}_{\mathrm{do}(X = x')}}(Y).
$$

In essence, by keeping one latent variable fixed and keeping the rest of the model unchanged (the $\mathrm{do}(\cdot)$ operator). By the assumption that our data distribution follows an SCM, which can be implicitly learned in a probabilistic data modeling proxy, such as a generative model, causal questions can be asked.

We will henceforth assume that our data distributions follow an (unknown) underlying SCM. Thus, the SCM data generation process can be implicitly learned in a probabilistic data modeling proxy, in our case, a flow matching model. Ultimately, if one can (partially) recover the SCM, the causal structure of the model (and, by extension, the data) can be explored and understood via the flow matching model.

\section{Background} \label{sec:background}
In this section, we introduce our main findings and the connection of linear maps with flow matching. In particular, we discuss how small perturbations in the latent space translate to changes in the data space via the Jacobian.

\subsection{Conditioning on Linear Interpolation of Linear Maps}
Let $\mathbf z, \tilde{\mathbf z} \sim \mathcal{N}(0, \mathbf{I}_d)$ where $\mathbf z$ and $\tilde{\mathbf z}$ are independent.
Let $\mathbf x = \mathbf A \tilde{\mathbf z}$ and define the interpolated vector as,

$$
\mathbf x_t = (1 - t) \mathbf z + t \mathbf x = (1 - t) \mathbf z + t \mathbf A \tilde{\mathbf z}.
$$

The pair $(\mathbf{x}, \mathbf{x}_t)$ is jointly Gaussian, so the conditional mean admits a closed form (derivation in Appendix~\ref{app:derivations}).
In particular,

$$
\mathbb{E}[\mathbf{x} \mid \mathbf{x}_t] = t \mathbf{A} \mathbf{A}^T \left( (1 - t)^2 \mathbf{I}_d + t^2 \mathbf{A} \mathbf{A}^T \right)^{-1}\mathbf{x}_t.
$$

At the extremes $t = 0$ and $t = 1$, the conditional mean simplifies to $\mathbb{E}[\mathbf{x} \mid \mathbf{x}_0] = 0$ and $\mathbb{E}[\mathbf{x} \mid \mathbf{x}_1] = \mathbf{x}_t$, as expected.

\subsubsection{Conditioning on arbitrary Gaussian}
For $\mathbf{x}_0 \sim \mathcal{N}(0, \mathbf{I}_d)$ and $\mathbf{x}_1 \sim \mathcal{N}(0, \boldsymbol \Sigma)$, the conditional mean again has a closed form (Appendix~\ref{app:derivations}). Returning to flow matching (see Equation~\eqref{eq:optimal_flow}), if one assumes that our data distribution $\mathcal{D} \sim p_{\text{data}}$ is (approximately) Gaussian, then,

\begin{equation*}
\mathbf{v}(\mathbf{x}_t,t) % = \mathbb{E}[\mathbf{x}_1 - \mathbf{x}_0 | \mathbf{x}_t] 
= (t \boldsymbol \Sigma - (1 - t) \mathbf{I}_d)[(1 - t)^2 \mathbf{I}_d + t^2 \boldsymbol \Sigma]^{-1} \mathbf{x}_t.
\end{equation*}

Thus, the flow matching ODE is,

\begin{align*}
\frac{d\mathbf{x}(t)}{dt} & = \mathbf{v}(\mathbf{x}_t, t) \\
& = \mathbf{A}(t) \mathbf{x}_t,
\end{align*}

where $\mathbf{A}(t) = (t \boldsymbol \Sigma - (1 - t) \mathbf{I}_d)[(1 - t)^2 \mathbf{I}_d + t^2 \boldsymbol \Sigma]^{-1}$.

\subsubsection{Conditioning beyond simple Gaussians}
We now study beyond simple Gaussians distributions and derive how we can express and (again) connect back to flow matching using a mixture of Gaussians instead (derivation in Appendix~\ref{app:derivations}). Let,

$$
\mathbf{x}_0 \sim \mathcal N(\boldsymbol \mu_0, \boldsymbol \Sigma_0), \quad \mathbf{x}_1 \sim \sum_{k=1}^K \pi_k \ \mathcal N(\boldsymbol \mu_k, \boldsymbol \Sigma_k),
$$

with independent $\mathbf{x}_0, \mathbf{x}_1$, mixture weights $\pi_k > 0$, $\sum_k \pi_k = 1$, and $t \in [0,1]$. We again define the linear interpolation,

$$
\mathbf{x}_t = (1-t) \mathbf{x}_0 + t \mathbf{x}_1.
$$

Conditioning on the latent component $K=k$ yields,

$$
\mathbf{x}_t \mid (K=k) \sim \mathcal N\left((1-t) \boldsymbol \mu_0 + t \boldsymbol \mu_k, (1-t)^2 \boldsymbol \Sigma_0 + t^2 \boldsymbol \Sigma_k \right).
$$

By marginalizing over $K$ we get the time density,

$$
p_t(\mathbf{x}) = \sum_{k=1}^K \pi_k \ \mathcal N\left(\mathbf{x} | (1-t) \boldsymbol \mu_0 + t \boldsymbol \mu_k, (1-t)^2 \boldsymbol \Sigma_0 + t^2 \boldsymbol \Sigma_k \right).
$$

Let $\mathbf{x} = (\mathbf{x}_0, \mathbf{x}_1)$ and introduce an unobserved component label $K \in \{1, \ldots, K \}$. We know by Bayes' rule that, \vspace{-1em}

\begin{equation}
\label{eq:mixture_responsibilities}
\begin{aligned}
\gamma_k(\mathbf{x}_t,t) & \coloneqq Pr(K = k \mid \mathbf{x}_t) \\
& = \frac{\pi_k \mathcal{N}(\mathbf{x}_t \mid \boldsymbol{\mu}_{t,k}, \boldsymbol{\Sigma}_{t,k})}{\sum_{\ell=1}^K \pi_\ell \mathcal{N}(\mathbf{x}_t \mid \boldsymbol{\mu}_{t,\ell}, \boldsymbol{\Sigma}_{t,\ell})},
\end{aligned}
\end{equation}

where $\boldsymbol{\mu}_{t,k} = (1-t) \boldsymbol{\mu}_0 + t \boldsymbol{\mu}_k$ and $\boldsymbol{\Sigma}_{t,k} = (1-t)^2 \boldsymbol{\Sigma}_0 + t^2 \boldsymbol{\Sigma}_k$.

Using conditional Gaussian identities (Appendix~\ref{app:derivations}), we obtain the mixture drift by marginalizing over $K$:

\begin{equation}
\label{eq:mixture_gaussian}
\begin{aligned}
\mathbf{v}(\mathbf{x}_t,t) 
& = \mathbb{E}[\mathbf{x}_1 - \mathbf{x}_0 \mid \mathbf{x}_t] \\
& = \sum_{k=1}^K \gamma_k(\mathbf{x}_t,t) \bigl[
    \boldsymbol{\mu}_k - \boldsymbol{\mu}_0 
    + \mathbf{A}_k(t) (\mathbf{x}_t - \boldsymbol{\mu}_{t,k})
\bigr],
\end{aligned}
\end{equation}

where $\boldsymbol{\mu}_{t,k} = (1-t) \boldsymbol{\mu}_0 + t \boldsymbol{\mu}_k$ and

\begin{equation*}
\mathbf{A}_k(t) = \left(t \boldsymbol{\Sigma}_k - (1-t) \boldsymbol{\Sigma}_0\right) \left((1-t)^2 \boldsymbol{\Sigma}_0 + t^2 \boldsymbol{\Sigma}_k \right)^{-1}.
\end{equation*}

\subsection{Findings on Gaussian Conditioning and Connection to Causality}
\label{subsec:gaussian_condition}

\begin{table*}[h]
    \centering
    \caption{Local Jacobians in three generative regimes. Intervention: we nudge the latent while holding the component label $K$ fixed. Observation: we only watch $\mathbf x_t$, so the posterior responsibilities $\gamma_k$ vary with $\mathbf x_t$.}
    \label{tab:jacobians}
    \resizebox{\textwidth}{!}{
    \begin{tabular}{l c c c@{\hspace{1em}}}
        \toprule
        \textbf{Scenario} & \textbf{Structural map ("cause $\to$ effect")}
        & \textbf{Jacobian under intervention} & \textbf{Jacobian under observation} \\
        \toprule
        \\
        Linear $\mathbf A$ & $\mathbf x = \mathbf A\mathbf z$
        & Constant $\mathbf A$ & Constant $\mathbf A$ \\
        \midrule
        \\
        Single--Gaussian flow & $\mathbf v(\mathbf x_t,t)=\mathbf A(t) \mathbf x_t+\mathbf b(t)$
        & Constant in $\mathbf x_t$ & The same affine map \\
        \midrule
        \\
        Mixture of Gaussians & Pick $K \sim \pi$ $\Longrightarrow \mathbf A_K(t) \mathbf x_t + \mathbf b_K(t)$
        & $\displaystyle \mathbf J_{\text{do}}(\mathbf x_t)=\mathbf A_K(t)$
        & Smooth but nonlinear, see Equation~\eqref{eq:jacobian-mixture} \\
        \bottomrule
    \end{tabular}
    }
\end{table*}

Table~\ref{tab:jacobians} summarizes how the local Jacobian, hence the outcome of the perturbation analysis, depends on the probabilistic structure that generates $\mathbf x_t$. When causal questions are asked (``\emph{do}'' notation), the latent component label $K$ is held fixed, hence the Jacobian is piecewise constant. From a purely observational standpoint the weights $\gamma_k(\mathbf x_t,t)$ depend on $\mathbf x_t$. The gradient of these weights is what bends the otherwise affine field. With $\mathbf x_0 \sim \mathcal N(0, \mathbf{I}_d)$ and $\mathbf x_1 \sim \mathcal N(0, \boldsymbol\Sigma)$, the flow matching drift,

\begin{equation*}
    \mathbf v(\mathbf x_t,t)=\bigl(t\boldsymbol\Sigma-(1-t)\mathbf I_d\bigr)
    \bigl[(1-t)^2\mathbf I_d+t^2\boldsymbol\Sigma\bigr]^{-1}    \mathbf x_t
\end{equation*}

is \emph{linear in $\mathbf x_t$}, so the Jacobian does not depend on $\mathbf x_t$.  Perturbation analysis is therefore trivial, $\Delta \dot{\mathbf x}_t = \mathbf A(t) \Delta \mathbf x_t$ and $\operatorname{Cov}(\Delta \mathbf x_t) = \mathbf A(t) \mathbf A(t)^{T}$ match the push-forward covariance exactly. Differentiating Equation~\eqref{eq:mixture_gaussian} with respect to $\mathbf x_t$ and applying the product rule gives,

\begin{equation}
    \nabla_{\mathbf x_t} \mathbf v = \sum_{k} \gamma_k \mathbf A_k(t) + \sum_{k}\left(\nabla_{\mathbf x_t} \gamma_k\right)
    \left[\boldsymbol \mu_k - \boldsymbol \mu_0 + \mathbf A_k(t) \Delta_k \right]
    \label{eq:jacobian-mixture}
\end{equation}

The first term is a convex blend of the affine maps $\mathbf A_k(t)$.  The second term injects the non-linearity. Since the gradient of a softmax (see Equation~\eqref{eq:mixture_responsibilities}) of quadratic forms can be written explicitly as,

\begin{equation*}
    \nabla_{\mathbf x_t} \gamma_k = \gamma_k \left(\mathbf S_k^{-1} (\mathbf m_k - \mathbf x_t) - \sum_j \gamma_j \mathbf S_j^{-1} (\mathbf m_j - \mathbf x_t)\right),
\end{equation*}

with $ \mathbf m_k = (1 -  t) \boldsymbol \mu_0 + t \boldsymbol \mu_k$ and $\mathbf S_k = (1-t)^2 \boldsymbol \Sigma_0 + t^2 \boldsymbol \Sigma_k$.  

Inside regions where a single $\gamma_k \approx1$ (i.e., a single component is dominant) the field is locally affine, non-linearity appears only in the transitions between components. To summarize our theoretical findings on the mixture of Gaussians case,

\begin{itemize}
    \itemsep0em
    \item \textbf{Mixture of affine drifts.}  Even though the global field can be nonlinear, it can be decomposed into a convex blend of $K$ affine rules.
    
    \item \textbf{Local linearity.}  Whenever a responsibility $\gamma_k$ dominates, the drift is effectively the $k$-th linear map.
    
    \item \textbf{Shared covariances.} If all $\boldsymbol \Sigma_k = \boldsymbol \Sigma$, then $\mathbf A_k(t) \equiv \mathbf A(t)$ and only the bias term jumps across components.
    
    \item \textbf{Endpoints.} At $t=0$ the drift reduces to $\sum_k \pi_k \boldsymbol \mu_k - \mathbf x_0$, at $t=1$ it equals $\mathbf x_1 - \boldsymbol \mu_0$, matching the prescribed boundary conditions $\mathbb{E}[\mathbf x_1] - \mathbf x_0$ and $\mathbf x_1 - \mathbb{E}[\mathbf x_0]$ respectively.
\end{itemize}

In conclusion, in a simple Gaussian setting, flow matching behaves (approximately) like a linear map. In the harder mixture case, the Jacobian remains tractable, it is the sum of a convex average of linear maps and a softmax‑gradient correction term. In other words, flow matching models learn approximately linear drifts whenever the underlying data distribution is itself close to Gaussian.

\section{Method} 
\label{sec:method}
In this section, we first introduce our methods to apply our findings, then connect them to diffeomorphic flow structure in the Gaussian setting and to causal language, and finally describe the models and datasets used in experiments.

\subsection{Wiggling in Flow Matching Models}
\label{subsec:Wiggle in models}

Given the foundations laid out so far, it is only natural to study the effect of small perturbations in the starting point for flow matching models and see the effects. Let $\mathbf{f}_\theta: \mathbb{R}^d \mapsto \mathbb{R}^d$ denote a (trained) flow matching model that maps a sample $\mathbf{x}_0 \sim p_{\text{init}}$ to a generated sample $\mathbf{\hat{x}_1} = \mathbf{f}_\theta(\mathbf{x}_0) \sim p_{\text{data}}$ by integrating the learned velocity field $\mathbf{v}_\theta$. By Taylor expansion we know that,

$$
\mathbf{f}_{\theta}(\mathbf{x}_0 + \varepsilon \mathbf{z}) \approx \mathbf{f}_{\theta}(\mathbf{x}_0) + \varepsilon \mathbf{J}_{\mathbf{f}_{\theta}}(\mathbf{x}_0) \mathbf{z}, \quad \mathbf{z} \sim \mathcal{N}(0, \mathbf{I}_d), \ \varepsilon > 0
$$

where,
$$
\mathbf{J}_{\mathbf{f}_{\theta}}(\mathbf{x}_0) = \frac{\partial}{\partial \mathbf {x}_0} \mathbf{f}_{\theta}(\mathbf{x}_0)
$$

is the Jacobian with respect to the input. 

Equivalently, the finite‐difference approximation can be used,

\begin{equation}
    \label{eq:jacobian_approx}
    \mathbf{J}_{\mathbf{f}_{\theta}}(\mathbf{x}_0) \mathbf{z} \approx \frac{\mathbf{f}_{\theta}(\mathbf{x}_0 + \varepsilon \mathbf{z}) - \mathbf{f}_{\theta}(\mathbf{x}_0)}{\varepsilon},
\end{equation}

which is a Jacobian-vector product for the random direction $\mathbf{z}$.

In practice, we use finite differences with small $\varepsilon$ and validate the stability of the estimate by comparing to automatic-differentiation JVPs and by sweeping $\varepsilon$ (Appendix~\\ref{app:diagnostics}).

\subsection{Connection to Diffeomorphism}

Consider the flow that transports $\mathbf x_0 \sim \mathcal{N}(\boldsymbol m_0, \boldsymbol \Sigma_0)$ to $\mathbf x_1 \sim \mathcal{N}(\boldsymbol m_1, \boldsymbol \Sigma_1)$. In this linear-Gaussian setting the flow map is explicit. Let,
\vspace{-0.5em}
\begin{align*}
\mathbf A & = \boldsymbol \Sigma_0^{-1/2}\left(\boldsymbol\Sigma_0^{1/2}\boldsymbol \Sigma_1 \boldsymbol \Sigma_0^{1/2}\right)^{1/2} \boldsymbol \Sigma_0^{-1/2}, \\
\mathbf L & = \log(\mathbf A), \\
\boldsymbol m(t) & = (1-t)\boldsymbol m_0 + t \boldsymbol m_1.
\end{align*}

Consider the flow map $\phi_t : \mathbf x_0 \mapsto \mathbf x(\mathbf x_0; t)$ with $\phi_t(\mathbf x_0) = \boldsymbol m(t) + e^{t\mathbf L} (\mathbf x_0 - \boldsymbol m_0).$ This solves the linear ODE,

$$
\dot{\mathbf x} = \mathbf L\left(\mathbf x - \boldsymbol m(t)\right) + \dot{\boldsymbol m}(t), \quad \mathbf x(0)=\mathbf x_0,
$$

hence $\phi_t$ is a diffeomorphism for all $t \in [0,1]$ in this construction, with derivative (Jacobian),

$$
Y(t) = \frac{\partial \phi_t}{\partial \mathbf x_0} = e^{t\mathbf L}.
$$

Pushing forward $\mathcal{N}(\boldsymbol m_0,\boldsymbol\Sigma_0)$ along $\phi_t$ gives,

$$
\mathbf x_t \sim \mathcal N \left(\boldsymbol m(t),\boldsymbol\Sigma(t)\right),
\quad
\boldsymbol \Sigma(t) = Y(t) \boldsymbol \Sigma_0 Y(t)^T.
$$

At $t=1$, $\phi_1(\mathbf x_0)=\boldsymbol m_1+\mathbf A(\mathbf x_0-\boldsymbol m_0)$ and $\boldsymbol\Sigma(1)=\boldsymbol\Sigma_1$, as required. For the flow $\phi_t$ above, finite differences recover the Jacobian exactly,

$$
\frac{\phi_t(\mathbf x_0 + \varepsilon \mathbf z) - \phi_t(\mathbf x_0)}{\varepsilon}
\xrightarrow[\varepsilon\downarrow 0]{}
Y(t) \mathbf z,
\quad
Y(t) = e^{t \mathbf L}.
$$

Consequently, the "wiggle" covariance propagated by $\phi_t$ matches the push‐forward covariance,

\begin{align*}
\operatorname{Cov}\left(\phi_t(\mathbf x_0 + \varepsilon \mathbf z) - \phi_t(\mathbf x_0)\right) & = \varepsilon^2 Y(t) \operatorname{Cov}(\mathbf z) Y(t)^T \\
& = \varepsilon^2 Y(t) Y(t)^T, 
\end{align*}

which is the empirical counterpart of the theoretical $\boldsymbol \Sigma(t) = Y(t) \boldsymbol \Sigma_0 Y(t)^T$ (with $\boldsymbol \Sigma_0 = \mathbf I_d$ in the standard initialization). This places the finite-difference Jacobian-vector product in direct correspondence with our conditioning formulas and validates the perturbation analysis in the Gaussian case.

\subsection{Discovering Attribute Relations via Perturbations}
\label{sec:jvp_attr}
So far we have discussed Jacobian-vector products for flow models in pixel space. However, for perceptual data, pixel-level dependencies are rarely meaningful for causal questions. In these cases, one needs a classifier to map generated images to semantic attributes. The pixel-space JVP only captures pixel correlations, not attribute-level dependencies. Thus, the attribute-level Jacobian-vector product becomes relevant to study.
We denote a (well-trained) classifier model $\mathbf{C}_{\theta} : \mathbb{R}^d \mapsto \mathbb{R}^n$, where $n$ denotes the number of attributes (we use logits in our experiments). Similar to Equation~\eqref{eq:jacobian_approx}, we can compute the Jacobian-vector product for the composed classifier-flow map as,
$$
\mathbf{J}_{\mathbf{C}_\theta \circ \mathbf{f}_\theta}(\mathbf{x}_0) \mathbf{z} \approx  \frac{\mathbf{C}_{\theta}(\mathbf{f}_\theta(\mathbf{x}_0 + \varepsilon\mathbf{z})) - \mathbf{C}_{\theta}(\mathbf{f}_\theta(\mathbf{x}_0))}{\varepsilon}, \quad \mathbf{z}\sim\mathcal{N}(0,\mathbf{I}_d).
$$
Let $\Delta \mathbf{y}(\mathbf{x}_0,\mathbf{z})$ denote the right-hand side above. By aggregating over many samples we estimate the second-moment matrix,

$$
\mathbf{S}(\mathcal{D}) = \mathbb{E}_{\mathbf{x}_0,\mathbf{z}} \left[\Delta \mathbf{y} \, \Delta \mathbf{y}^T \right].
$$

When we report a correlation matrix, we use the normalized version
$
\mathrm{Corr}(\mathcal{D}) = \mathbf{D}^{-1/2}\mathbf{S}(\mathcal{D})\mathbf{D}^{-1/2},
$
where $\mathbf{D}=\mathrm{diag}(\mathbf{S}(\mathcal{D}))$.

\subsection{Models \& Architectures}
\label{subsec:Models}

\begin{table}[h]
  \centering
  \caption{Architectural configurations of the UNet-based models.}
  \label{tab:models}
  \resizebox{\columnwidth}{!}{
  \begin{tabular}{lcccccc}
    \toprule
    \textbf{Model} & \textbf{Input} & \textbf{Output} & \textbf{Time Emb. Dim} & \textbf{Base Ch.} & \textbf{Channel Mult.} & \textbf{Attention Res.} \\
    \midrule
    FlowUNet (baseline)  
      & $C = 1$ 
      & $C = 1$ 
      & 256 
      & 64 
      & (1, 2, 4, 8, 16) 
      & – \\
    \quad \textit{(bilinear up?)} 
      & – & – & – & \multicolumn{1}{c}{yes / no} & – & – \\
    \midrule
    FlowUNet (attention)   
      & $C = 3$ 
      & $C = 3$ 
      & 256 
      & 64 
      & (1, 1, 2, 2, 4) 
      & (8, 16, 32) \\
    \bottomrule
  \end{tabular}
  }
\end{table}

\begin{table}[H]
  \centering
  \caption{Training hyperparameters used in our experiments.}
  \label{tab:hyperparams}
  \begin{tabular}{lcc}
    \toprule
    \textbf{Hyperparameter}   & \textbf{Default / Value} \\
    \midrule
    Training steps           & 400\,001 \\
    Batch size               & 128 \\
    Learning rate (AdamW)    & $2\times10^{-4}$ \\
    \bottomrule
  \end{tabular}
\end{table}

Since the introduction of the UNet architecture by \cite{ronneberger2015}, many generative models have adapted this architecture. Thus, the UNet-like architecture was a natural choice for doing practical experiments. The tables above are the baseline architecture and hyperparameters that have been used in all experiments, and in the cases that deviate, we refer to the code \ref{app:Code}. For the classifier, a ResNet50 model \cite{resnet50} was trained and used in which the last layer was fine-tuned on the respective datasets. Both implementations of the architectures can be found in the code at \ref{app:Code}.
\subsection{Datasets and Experimental Setup}
\label{subsec:datasets}
We evaluate on synthetic distributions and standard image datasets. All experiments use PyTorch and the data loaders in our codebase; unless otherwise stated, $t \sim \mathcal{U}(0,1)$ and the interpolation is linear.

\textbf{Synthetic Gaussians.} We use 1D standard Gaussians for the closed-form comparison, and a 2D Gaussian with correlation $\rho=0.55$ for the MLP flow experiment. The 1D neural flow is trained on 10{,}000 samples with a small MLP (hidden size 128, learning rate $10^{-3}$, 2000 training steps). The 2D MLP uses 10 layers with width 1024 (learning rate $10^{-4}$, 1000 training steps). For the mixture-of-Gaussians analysis, we use the analytical drift/Jacobian formulas and compare against numerical estimates.

\textbf{MNIST.} For the pixelwise perturbation analysis we use MNIST digits and focus on the digit ``5'' subset. Images are $28\times 28$ grayscale with a simple \texttt{ToTensor} transform (no normalization). The flow model is a UNet-based architecture (Table~\ref{tab:models}) trained with batch size 64 and learning rate $2\times10^{-4}$ for 10{,}001 steps. We estimate correlations using batches of generated samples and compare against empirical pixel correlations from the dataset.

\textbf{CelebA.} We use the aligned and cropped CelebA dataset with 40 binary attributes. Images are resized to $64\times64$ using a center crop and \texttt{ToTensor}. For attribute-level analysis we train a ResNet50 classifier with ImageNet initialization, replacing the final layer with 40 outputs and optimizing a BCE-with-logits loss (batch size 128, learning rate $2\times10^{-4}$, 20 epochs). We focus on the attributes \texttt{Heavy\_Makeup} (19), \texttt{Pointy\_Nose} (28), and \texttt{Rosy\_Cheeks} (30), and compute the empirical attribute correlation matrix from the provided labels as ground truth. Our flow model for CelebA uses the UNet-based flow architecture in Table~\ref{tab:models} (FlowUNet); we report results for 10{,}000 generated samples and a conditioned subset selected by a small Jacobian-norm threshold for \texttt{Heavy\_Makeup}. We also experimented with an attention UNet variant, but those results are not included in this paper.

\section{Results} \label{sec:result}
In this section, we provide our empirical results and findings on both toy datasets and standard image datasets. 

\subsection{Gaussian to Gaussian in \texorpdfstring{$\mathbb{R}$}{R}}
Firstly, we will begin with the simplest case to showcase our findings.

\begin{figure}[h]
    \centering
    \includegraphics[width=\columnwidth]{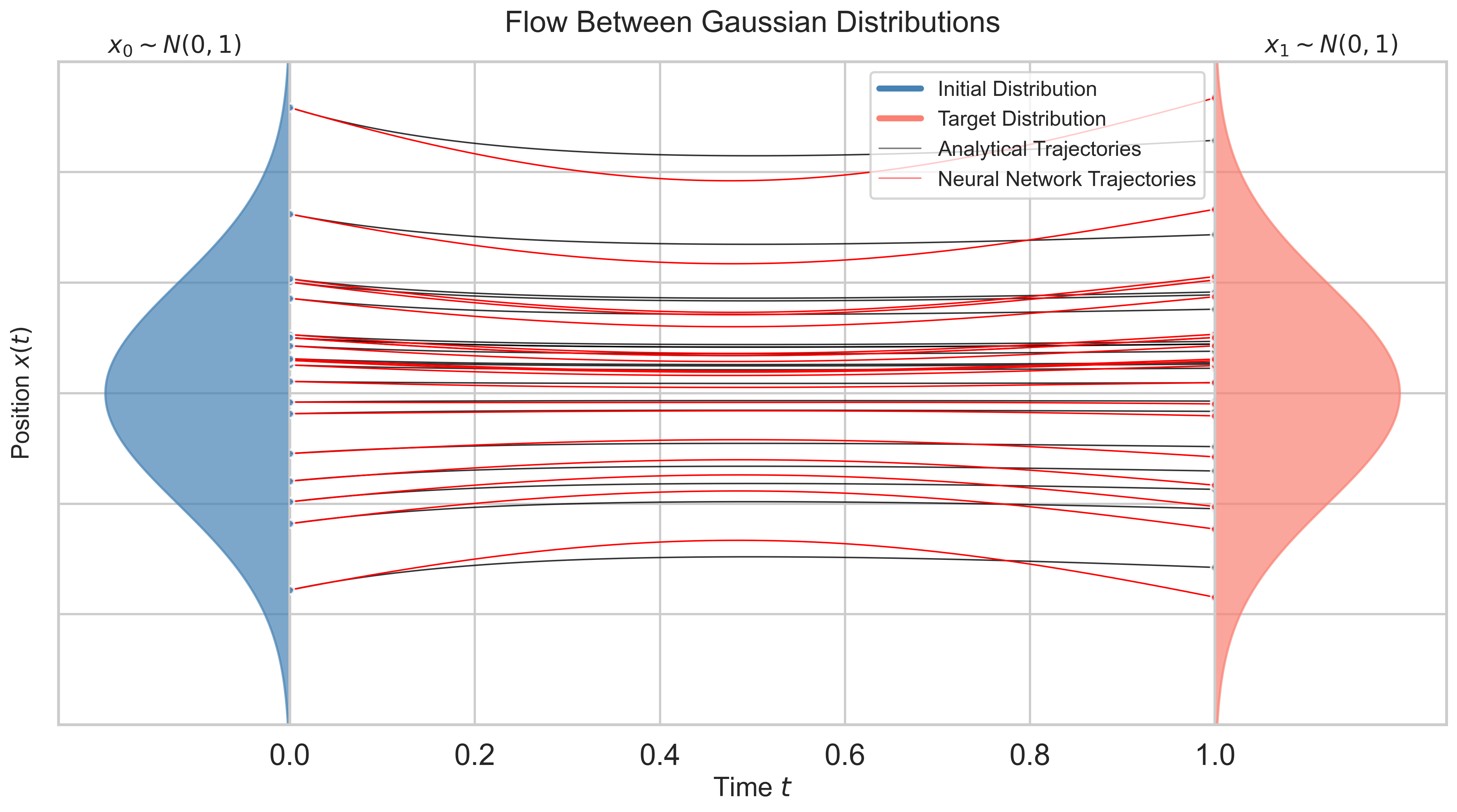}
    \caption{Flow Matching between two standard Gaussian distributions.}
    \label{fig:gaussian_1D_comparison}
\end{figure}

One can observe that in Figure~\ref{fig:gaussian_1D_comparison} it is apparent that the empirical results align with our theoretical work. One can see that there is a difference in the analytical trajectories (as proposed in \ref{sec:background}) and the learned ones. For this simple case, a simple MLP model was used instead of the UNet-architecture (see the Code~\ref{app:Code} for further details).

We want to emphasize an important aspect that may appear odd at first glance. It is a common misconception that flow matching learns straight trajectories. This is not true since the predictions are not for a single point but the average over a larger distribution. Thus, flowing straight between points is not equivalent to flowing straight between distributions.

\subsection{Mixture of Gaussians in \texorpdfstring{$\mathbb{R}^2$}{R2}}

Secondly, we will extend our simple example by a dimension and observe the more complex mixture of Gaussians case.

\begin{figure}[h]
     \centering
         \includegraphics[width=\columnwidth]{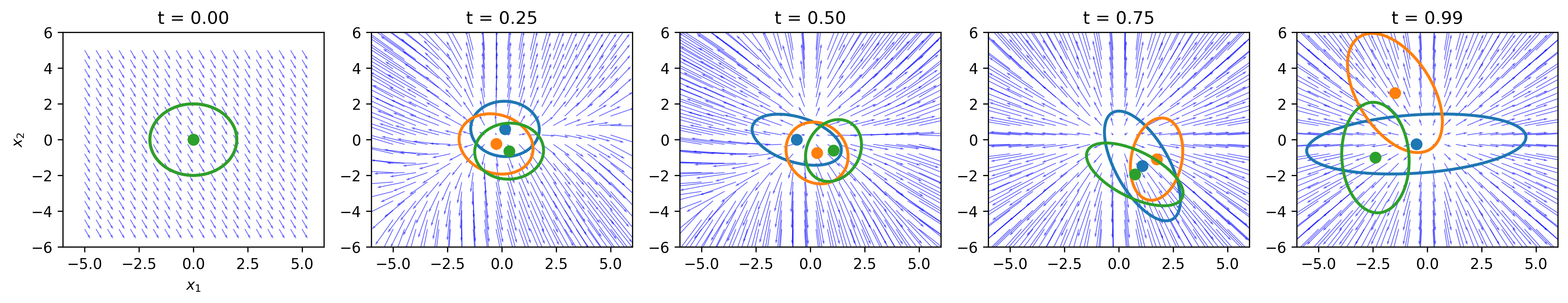}
        \caption{Time Evolution of Mixture of Gaussians Example.}
        \label{fig:mixture_of_gaussians}
\end{figure}

In Figure~\ref{fig:mixture_of_gaussians} we see the time evolution from a normal 2D Gaussian to a mixture of Gaussians (where $K = 3$). Further, here we will present our results,

\begin{table}[h]
    \centering
    \caption{Jacobian Analysis in Mixture of Gaussians}
    \resizebox{\columnwidth}{!}{
    \begin{tabular}{ccc}
        \toprule
        \textbf{Time $t$} & \textbf{Numerical vs. Analytical} & \textbf{Intervention vs. Analytical} \\
        \midrule
        0.00 & 0.058 & $<$0.001 \\
        0.25 & 0.289 & 0.284 \\
        0.50 & 0.219 & 0.300 \\
        0.75 & 0.251 & 0.039 \\
        0.99 & 0.179 & $<$0.001 \\
        \bottomrule
    \end{tabular}
    }
    \label{tab:mixture_gaussian_results}
\end{table}

Table~\ref{tab:mixture_gaussian_results} summarizes the average difference in methods versus ground truth analytical solutions. Again, our empirical results align with our theoretical claims, where (and when) the components are overlapping the difference of our methods becomes larger, but as time passes (or, if spatially it becomes "clearer") the methods yield a low difference. This suggests that if, both temporally and spatially, it is clear which component is dominating, the flow becomes a linear map.

\subsection{Pixel-Space Validation}
As discussed in \ref{sec:method}, our methods are applicable in pixel space. 

\begin{figure}[h]
    \centering
    \begin{subfigure}[b]{0.3\columnwidth}
        \centering
        \includegraphics[width=\columnwidth]{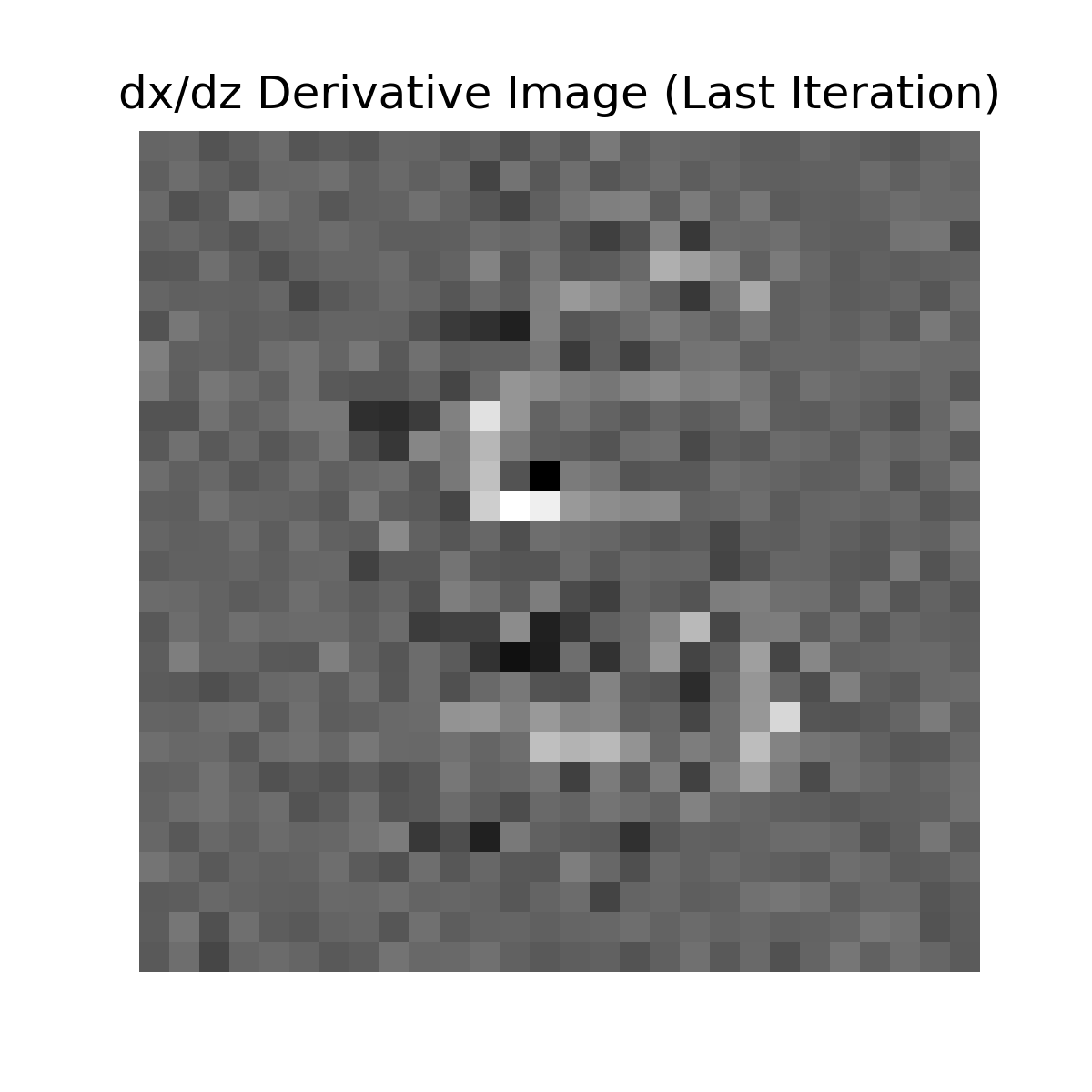}
        \caption{Directional derivative}
        \label{fig:dxdz}
    \end{subfigure}
    \begin{subfigure}[b]{0.3\columnwidth}
        \centering
        \includegraphics[width=\columnwidth]{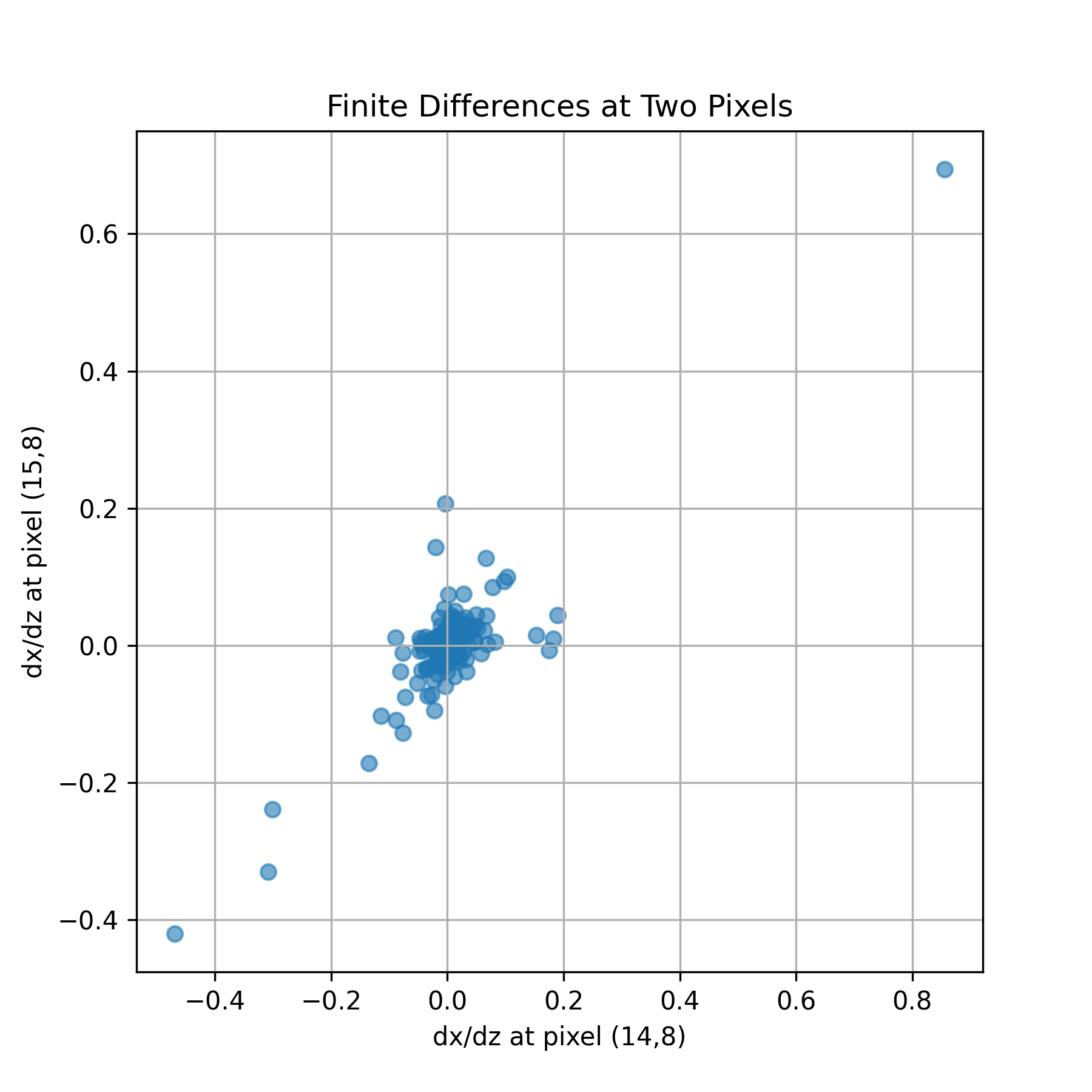}
        \caption{Finite Difference}
        \label{fig:dxdzdiff}
    \end{subfigure}
        \begin{subfigure}[b]{0.3\columnwidth}
        \centering
        \includegraphics[width=\columnwidth]{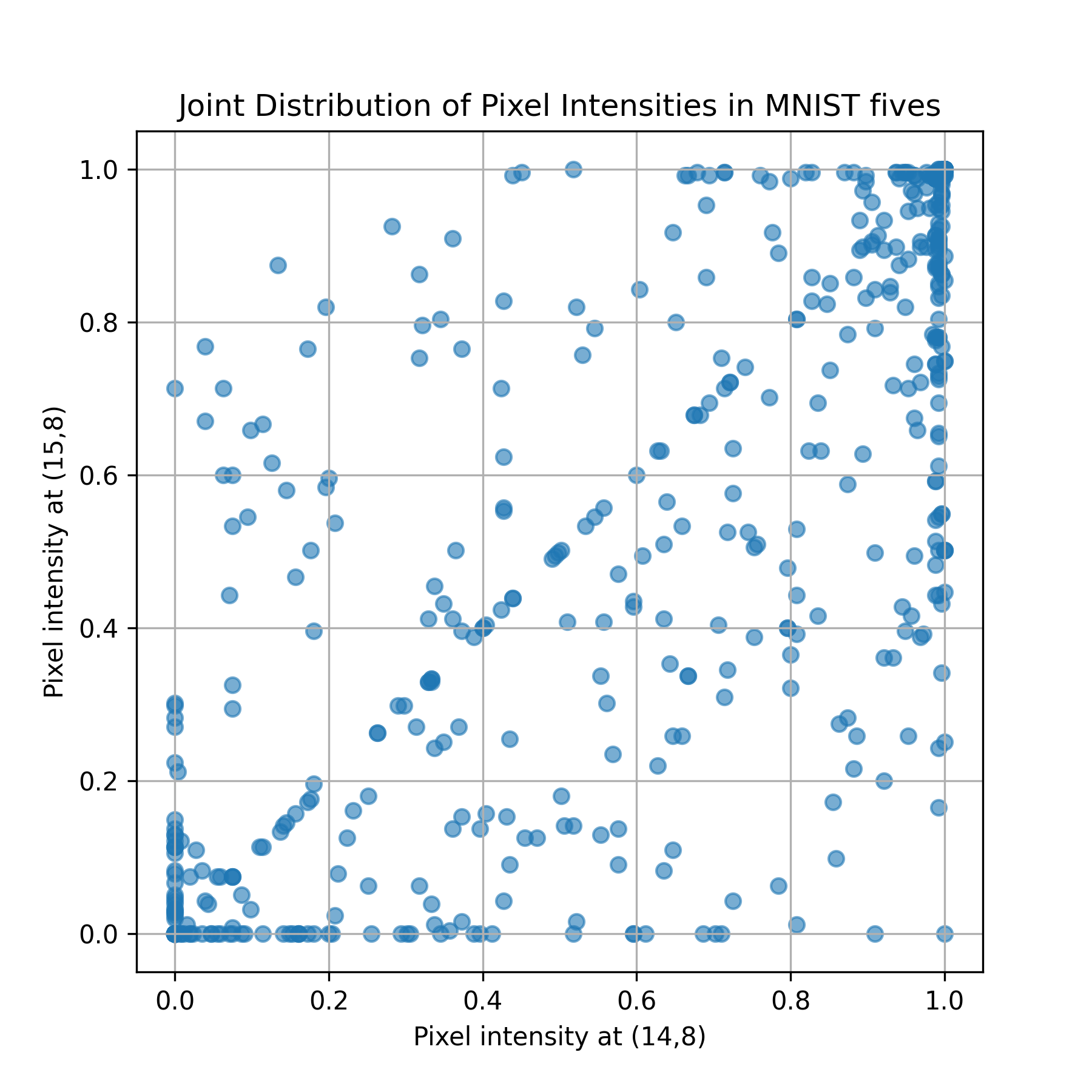}
        \caption{True joint distribution}
        \label{fig:jointdistfives}
    \end{subfigure}
    \caption{(Left) is the approximate "derivative image" of an MNIST five; (Middle) is the finite difference of pixel pairs over MNIST fives; (Right) is the true join distribution of the pixel pairs.}
    \label{fig:pixelwise}
\end{figure}

\begin{table}[h]
    \centering
    \caption{Correlation Coefficient Matrices}
    \resizebox{\columnwidth}{!}{
    \begin{tabular}{cc}
        \toprule
        \textbf{Estimated (ours)} & \textbf{Empirical (ground truth)} \\
        \midrule
        $
        \mathrm{Corr}(\mathcal{D}_5) = 
        \begin{bmatrix}
        1 & 0.8674 \\
        0.8674 & 1
        \end{bmatrix} 
        $ 
        & 
        $
        \mathrm{Corr}(\mathcal{D}_5) = 
        \begin{bmatrix}
        1 & 0.8813 \\
        0.8813 & 1
        \end{bmatrix} 
        $ \\
        \bottomrule
    \end{tabular}
    }
    \label{tab:ccm}
\end{table}

From Figure~\ref{fig:pixelwise} we can summarize the pixel-space validation. In this experimental setting, we focus on the fives (5) in the MNIST dataset \cite{deng2012mnist} and on the pixels $(14, 8)$ and $(15, 8)$. Our JVP-based estimator recovers their empirical pairwise correlation (Table~\ref{tab:ccm}). We also observed similar behavior for larger pixel regions, but omit those results for brevity.

\subsection{Attribute-Level Analysis}
We now apply our perturbation-based analysis to a high-dimensional image dataset. CelebA is a face attribute dataset that contains more than $200 \ 000$ celebrity images and 40 binary attributes. A standardized alignment and cropped version were used, resized to $64 \times 64$ pixels \cite{CelebA}.

\begin{figure}[H]
    \centering
    \includegraphics[width=0.8\columnwidth]{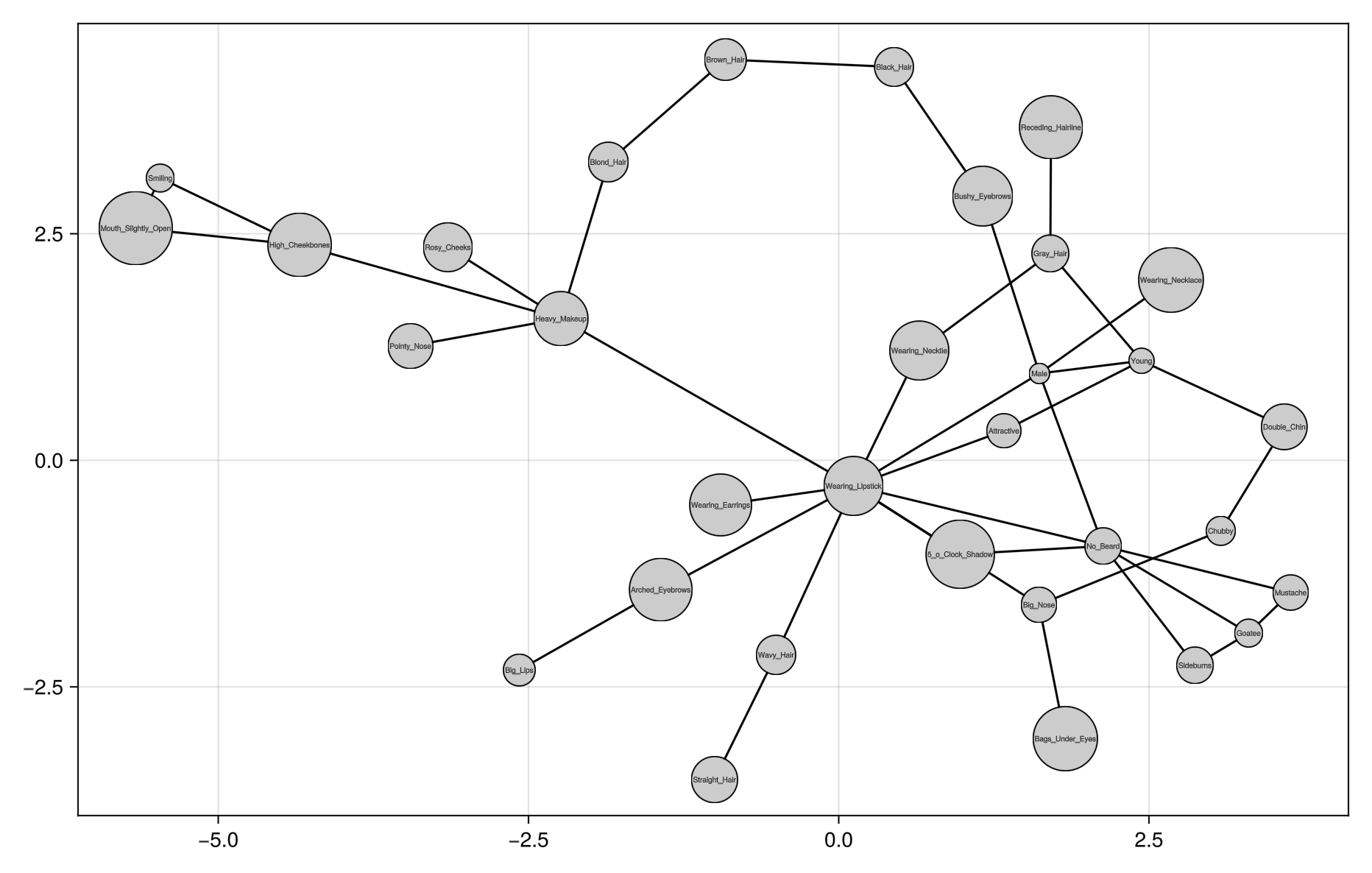}
    \caption{Estimated Conditional Dependence Structure of the CelebA dataset.}
    \label{fig:ecds_celeba}
\end{figure}

To form a hypothesis about attribute dependencies within the CelebA dataset, we first estimated a conditional dependence structure directly from the 40 binary attribute labels provided with the dataset. This analysis was performed using the \texttt{CausalInference.jl} package by \cite{Schauer_CausalInference_jl}, to infer the structure shown in Figure~\ref{fig:ecds_celeba}. This graph represents putative relationships between the attributes of the ground truth. For the experiments, a subset of features was selected. \texttt{Pointy\_Nose} (attribute 28), \texttt{Rosy\_Cheeks} (attribute 30), and \texttt{Heavy\_Makeup} (attribute 19) were selected. According to the estimated graph, attribute 19 serves as a putative common cause between the two other features, and therefore this subset is interesting to study. If attribute 19 truly confounds the relationship between attributes 28 and 30, then conditioning on its absence should reduce their correlation.

For the conditional analysis on the CelebA dataset, we selected samples where the JVP norm of the composed map for attribute 19, $\Vert \mathbf{J}^{19}_{\mathbf{C}_\theta \circ \mathbf{f}_\theta} \Vert$ (abbreviated as $\|\mathbf{J}^{19}\|$), was below a heuristic threshold of $10^{-4}$. This threshold was chosen to select samples in which small latent perturbations induce minimal change in the classifier's output for \texttt{Heavy\_Makeup}. We emphasize that this procedure constitutes \emph{conditioning} on a subset of the data, i.e., estimating $P(y \mid \|\mathbf{J}^{19}\| \text{ small})$, rather than a true $\mathrm{do}$-operation that would modify the data-generating process itself. Nevertheless, this allows for an initial exploration of whether attribute dependencies change in a manner consistent with the hypothesized causal structure. A more systematic investigation into the sensitivity of our findings to this threshold, and the development of principled methods for defining true structural interventions via the generative model, remain important directions for future research.

\begin{table}[H]
    \centering
    \caption{Correlation of Attributes (JVP-based estimate)}
    \resizebox{\columnwidth}{!}{
    \begin{tabular}{ccc}
        \toprule
        \textbf{Empirical (ground truth)} & \textbf{Estimated (ours)}  & \textbf{Estimated Conditional (ours)} \\
        \midrule
        200 000 samples & 10 000 samples & 1 700 samples \\
        \midrule
        $
        \mathrm{Corr}(\mathcal{D}_{28, 30}) \approx 0.40
        $ 
        & 
        $
        \mathrm{Corr}(\mathcal{D}_{28, 30}) \approx 0.16
        $ 
        &
        $
        \mathrm{Corr}(\mathcal{D}_{28, 30})_{\|\mathbf{J}^{19}\| \to 0} \approx 0.04        
        $
        \\
        \bottomrule
    \end{tabular}
    }
    \label{tab:corr_attr}
\end{table}

\begin{figure}[H]
    \centering
    \begin{subfigure}[t]{0.32\columnwidth}
        \centering
        \vspace{0pt}
        \includegraphics[width=\linewidth]{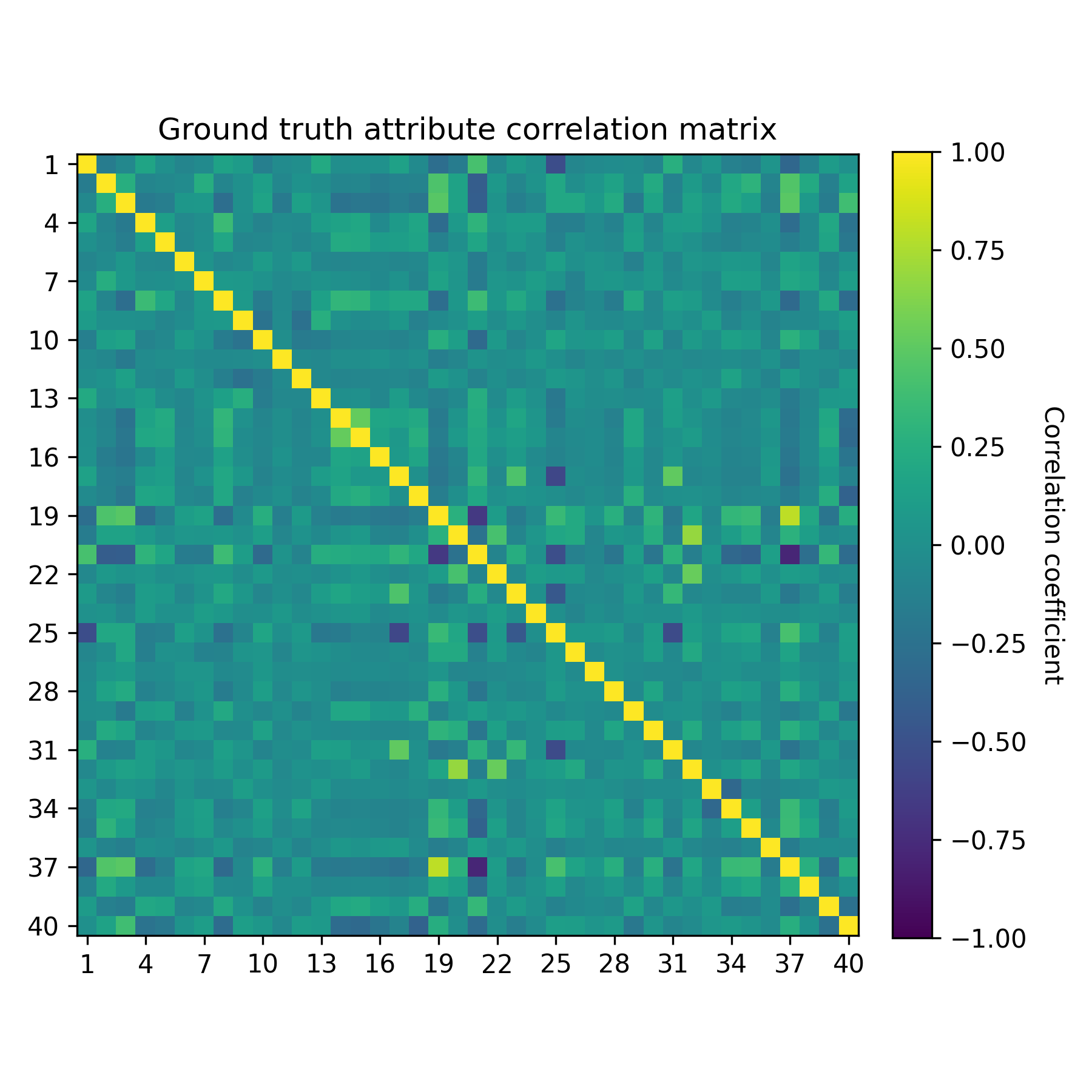}
        \caption{Ground truth correlation.}
        \label{fig:groundtruth}
    \end{subfigure}
    \hfill
    \begin{subfigure}[t]{0.32\columnwidth}
        \centering
        \vspace{0pt}
        \includegraphics[width=\linewidth]{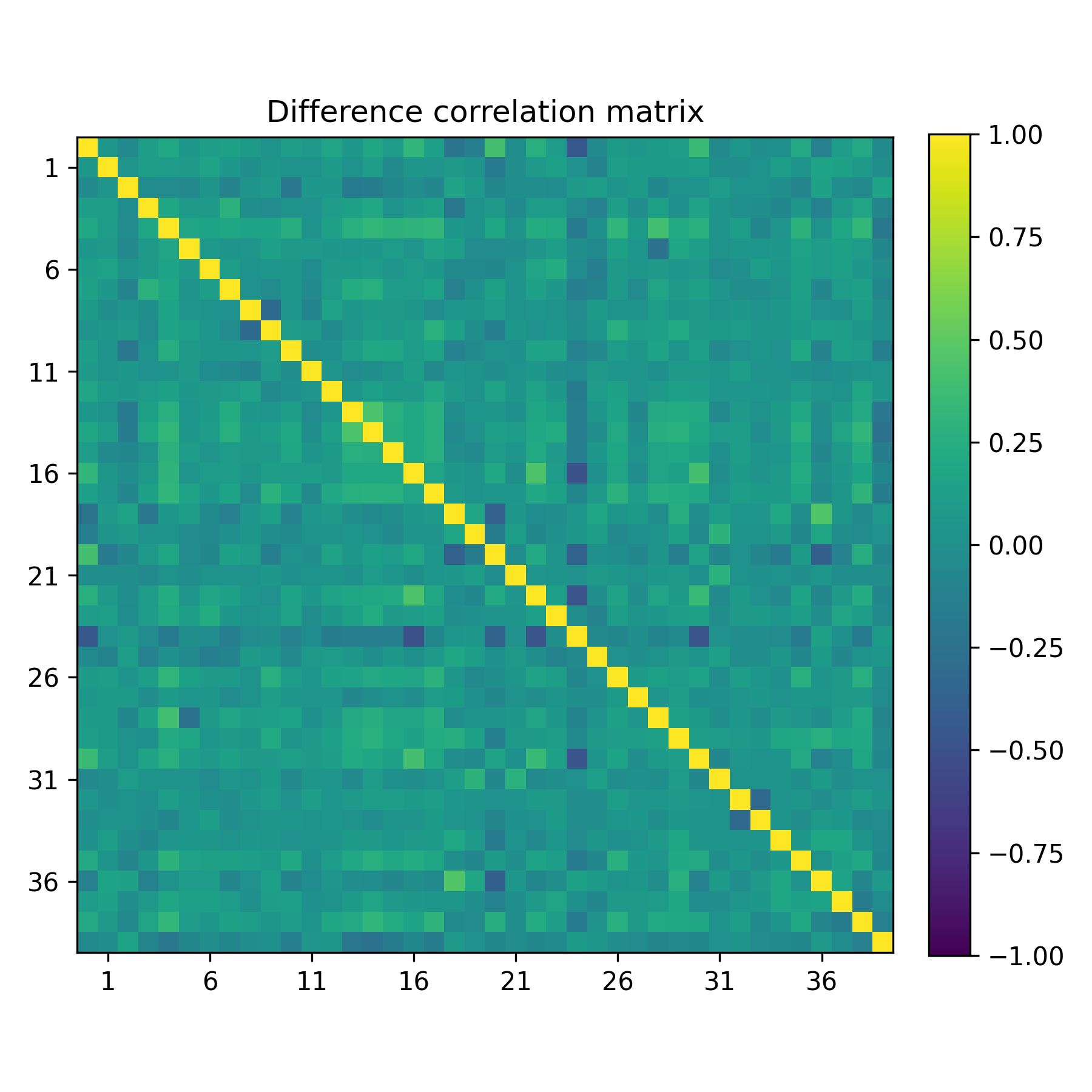}
        \caption{Estimated $\mathrm{Corr}(\mathcal{D})$ (JVP-based).}
        \label{fig:estimated_observation}
    \end{subfigure}
    \hfill
    \begin{subfigure}[t]{0.32\columnwidth}
        \centering
        \vspace{0pt}
        \includegraphics[width=\linewidth]{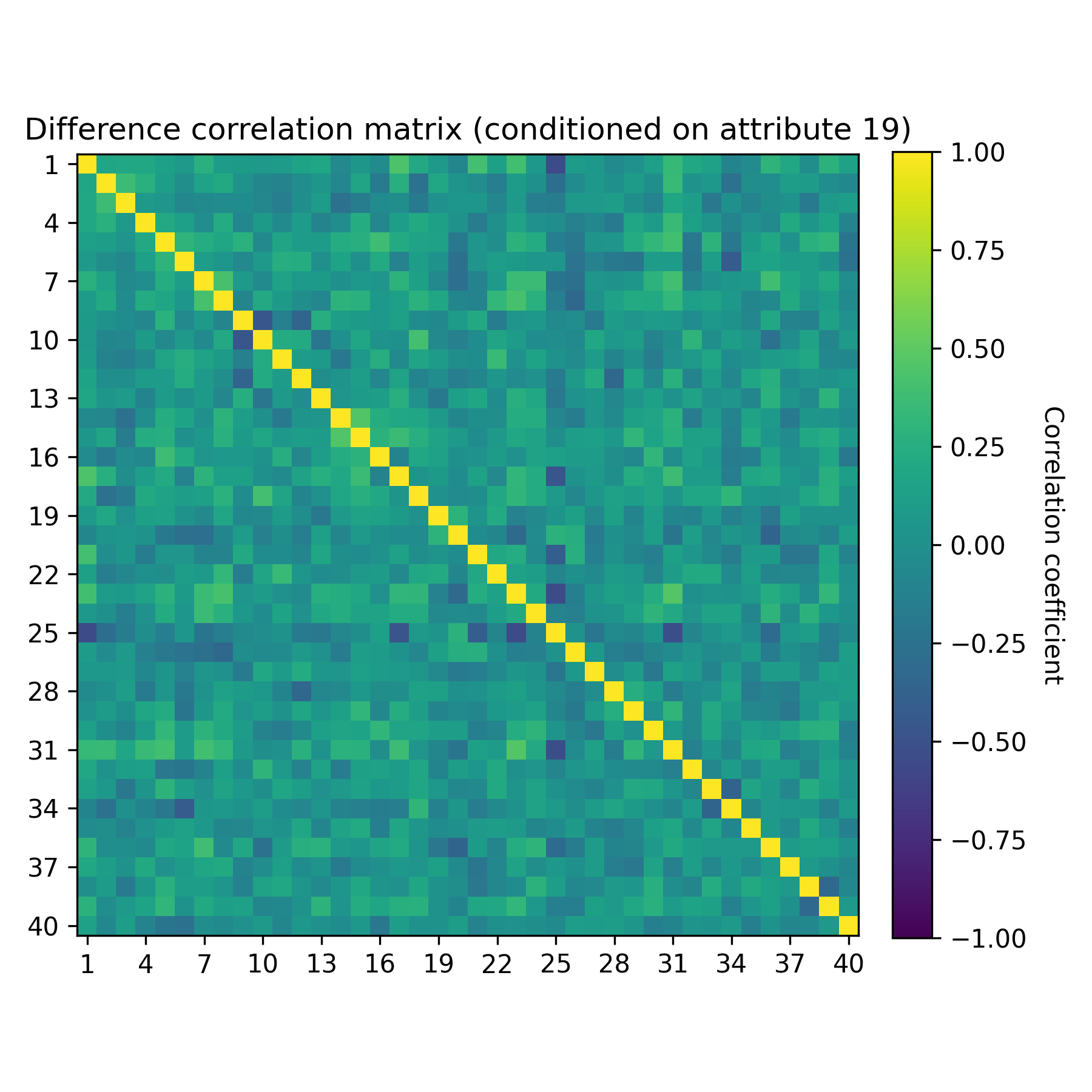}
        \caption{Estimated $\mathrm{Corr}(\mathcal{D})_{\|\mathbf{J}^{19}\| \to 0}$.}
        \label{fig:estimated_intervention}
    \end{subfigure}
    \caption{The empirical ground truth correlation matrix, the estimated correlation matrix under observation, and the estimated correlation matrix after conditioning on small $\|\mathbf{J}^{19}\|$.}
    \label{fig:estimated_corr_VS_ground_truth}
\end{figure}

Table~\ref{tab:corr_attr} summarizes our main findings for the pairwise correlation using the normalized JVP-based estimator from Section~\ref{sec:jvp_attr}. Using an order of magnitude fewer samples (10\,000 vs.\ 200\,000), our method partially recovers the empirical correlation, and we hypothesize that the estimate would improve with more samples. The conditioned subset is selected by the JVP criterion and contains roughly 1\,700 samples out of 10\,000. When we condition on samples with near-zero Jacobian norm for attribute 19, the correlation drops close to zero. This is \emph{consistent with} attribute 19 being a common cause for attributes 28 and 30, though it does not constitute definitive causal evidence, since conditioning on a common cause is expected to reduce downstream correlations under both causal and purely associational interpretations.

Figure~\ref{fig:groundtruth} depicts the empirical correlation matrix computed directly from the 40 binary attribute labels in the CelebA dataset, serving as a reference. Figure~\ref{fig:estimated_observation} shows the estimated attribute correlation matrix from our Jacobian-vector-product-based method, which is visually close to the ground truth. Figure~\ref{fig:estimated_intervention} shows the estimated correlation matrix after conditioning on small $\|\mathbf{J}^{19}\|$.

\section{Discussion and Conclusion}
\label{sec:discussion_conclusion}
In this work, we have established two main results. First, by deriving closed-form expressions for the conditional expectations in the Gaussian and mixture of Gaussians settings, we show that even when the global drift field is nonlinear, it admits local affine decompositions whose Jacobians are analytically tractable. Second, by analyzing how perturbations propagate through the learned flow, we demonstrate that the empirical Jacobian can recover dependency structures between generated features, both at the pixel level and, when composed with a classifier, at the attribute level.

We emphasize an important distinction: our empirical ``intervention'' in the CelebA experiments (conditioning on samples where the classifier Jacobian norm for a given attribute is small) constitutes conditioning on a subset of the data, i.e., $P(y \mid \|\mathbf{J}\| \text{ small})$, rather than a true $\mathrm{do}$-operation that modifies the data-generating process. While the results are \emph{consistent with} the hypothesized causal structure (the correlation between two attributes drops when we condition on their putative common cause), this does not constitute causal inference in the formal sense of Pearl's framework. Formalizing the connection between Jacobian-based filtering and structural interventions remains an open problem.

\subsection{Limitations}
Several limitations warrant consideration. First, our closed-form theoretical results assume Gaussian or mixture-of-Gaussian structure; real-world data distributions are considerably more complex, and the quality of the local affine approximation in such settings is not yet characterized. Second, relying on a pretrained classifier $\mathbf{C}_\theta$ to map pixel perturbations to semantic labels introduces inductive biases that may affect estimated dependencies. Third, while we compute Jacobian-vector products (not the full Jacobian matrix), the number of probes needed to estimate the correlation structure can still be costly for large-scale models. Fourth, the Jacobian-norm threshold used to approximate interventions is heuristic, and a sensitivity analysis of this choice has not been conducted. Finally, we have not tested the method on known causal structures beyond common causes; in particular, a collider test (where conditioning on a collider should \emph{increase} rather than decrease correlations) would provide an important sanity check.

\subsection{Future Directions}
Building on our findings and the feedback received, we identify several promising directions:

\begin{itemize}
  \itemsep0em
  \item \textbf{From conditioning to intervention}: Developing methods that modify the generative flow itself (rather than filtering its outputs) to implement true $\mathrm{do}$-operations via Jacobian manipulation.

  \item \textbf{Causal structure testing}: Evaluating the method on synthetic datasets with known causal graphs containing common causes, colliders, and mediators, to rigorously assess whether the approach can distinguish between these structures.

  \item \textbf{Beyond Gaussians}: Extending the local affine characterization to richer distribution families, or establishing conditions under which the Gaussian approximation remains adequate.

  \item \textbf{Counterfactual inference}: Extending to level 3 of the causal ladder, $P(y_{x'}\mid x,y)$, within flow matching inference.

  \item \textbf{Scalable JVP methods}: Leveraging automatic differentiation, randomized sketching, or power iteration to reduce the number of probes needed for high-dimensional settings.

  \item \textbf{Connections to disentanglement}: Relating our Jacobian-based analysis to the disentanglement literature~\cite{wei2021orthogonal, khrulkov2021disentangled}, where Jacobian regularization is used to encourage independent latent factors.
\end{itemize}

\subsection{Concluding Remarks}
In summary, we have provided a theoretical foundation for understanding how flow matching models encode local dependency structure, and an empirical methodology for extracting this structure via small latent perturbations. While the causal interpretation of our findings requires further development (particularly the formalization of true interventions within the generative process), we believe this work lays useful groundwork for more transparent and interpretable generative models.

\subsubsection*{Acknowledgements}
The computations and data handling were enabled by resources provided by the
\href{https://naiss.se}{National Academic Infrastructure for Supercomputing in Sweden (NAISS)},
via grant agreement no.\ NAISS 2024/22-1625 (NAISS Small Compute at C3SE), including NVIDIA A40 GPUs, partially funded by the Swedish Research Council.

\clearpage
\appendix
% \section{Appendix}
\section{Code}
\label{app:Code}
The source code and scripts for this project are released at \href{https://github.com/rezaarezvan/causdiff}{github.com/rezaarezvan/causdiff} under the MIT license.

\section{Derivations for Gaussian Conditioning}
\label{app:derivations}

\subsection{Linear Interpolation of Linear Maps}
Let $\mathbf z, \tilde{\mathbf z} \sim \mathcal{N}(0, \mathbf{I}_d)$ be independent, $\mathbf x = \mathbf A \tilde{\mathbf z}$, and $\mathbf x_t = (1-t)\mathbf z + t \mathbf x$. Then $(\mathbf x, \mathbf x_t)$ is jointly Gaussian with mean zero and covariance
\begin{equation*}
\boldsymbol \Sigma = \begin{bmatrix}
\mathbf A\mathbf A^T & t\mathbf A\mathbf A^T \\
 t\mathbf A\mathbf A^T & (1-t)^2 \mathbf I_d + t^2 \mathbf A\mathbf A^T
\end{bmatrix}.
\end{equation*}
For a zero-mean jointly Gaussian pair $(\mathbf u, \mathbf v)$ with blocks $\boldsymbol \Sigma_{\mathbf u\mathbf u}$ and $\boldsymbol \Sigma_{\mathbf v\mathbf v}$, the conditional mean is
\begin{equation*}
\mathbb{E}[\mathbf u\mid \mathbf v] = \boldsymbol \Sigma_{\mathbf u\mathbf v}\,\boldsymbol \Sigma_{\mathbf v\mathbf v}^{-1}\,\mathbf v.
\end{equation*}
Substituting yields
\begin{equation*}
\mathbb{E}[\mathbf x\mid \mathbf x_t] = t \mathbf A\mathbf A^T \left((1-t)^2 \mathbf I_d + t^2 \mathbf A\mathbf A^T\right)^{-1}\mathbf x_t.
\end{equation*}

\subsection{Arbitrary Gaussian Targets}
For $\mathbf{x}_0 \sim \mathcal{N}(0,\mathbf I_d)$ and $\mathbf{x}_1 \sim \mathcal{N}(0,\boldsymbol\Sigma)$, with $\mathbf{x} = \mathbf{x}_1-\mathbf{x}_0$ and $\mathbf{x}_t=(1-t)\mathbf{x}_0+t\mathbf{x}_1$, the same conditional Gaussian identity yields
\begin{equation*}
\mathbb{E}[\mathbf x\mid \mathbf x_t] = \mathrm{Cov}(\mathbf x,\mathbf x_t)\,\mathrm{Cov}(\mathbf x_t,\mathbf x_t)^{-1}\,\mathbf x_t,
\end{equation*}
which leads to the Gaussian drift used in the main text.

\subsection{Mixture-of-Gaussians Drift}
Let $\mathbf{x}_0 \sim \mathcal N(\boldsymbol\mu_0,\boldsymbol\Sigma_0)$ and $\mathbf{x}_1 \sim \sum_{k=1}^K \pi_k\,\mathcal N(\boldsymbol\mu_k,\boldsymbol\Sigma_k)$ with $\mathbf{x}_t=(1-t)\mathbf{x}_0+t\mathbf{x}_1$. The responsibilities
\begin{equation*}
\gamma_k(\mathbf x_t,t)=\frac{\pi_k\,\mathcal N(\mathbf x_t\mid \boldsymbol\mu_{t,k},\boldsymbol\Sigma_{t,k})}{\sum_{\ell}\pi_\ell\,\mathcal N(\mathbf x_t\mid \boldsymbol\mu_{t,\ell},\boldsymbol\Sigma_{t,\ell})}
\end{equation*}
with $\boldsymbol\mu_{t,k}=(1-t)\boldsymbol\mu_0+t\boldsymbol\mu_k$ and $\boldsymbol\Sigma_{t,k}=(1-t)^2\boldsymbol\Sigma_0+t^2\boldsymbol\Sigma_k$ follow from Bayes' rule. Conditioning on $K=k$ gives a jointly Gaussian pair $(\mathbf x_0,\mathbf x_1)$, and with $\mathbf S_k=(1-t)^2\boldsymbol\Sigma_0+t^2\boldsymbol\Sigma_k$ and $\boldsymbol\delta_k=\mathbf x_t-(1-t)\boldsymbol\mu_0-t\boldsymbol\mu_k$, we obtain
\begin{align*}
\mathbb{E}[\mathbf{x}_1 \mid \mathbf{x}_t, K=k] &= \boldsymbol\mu_k + t\boldsymbol\Sigma_k\,\mathbf S_k^{-1}\,\boldsymbol\delta_k, \\
\mathbb{E}[\mathbf{x}_0 \mid \mathbf{x}_t, K=k] &= \boldsymbol\mu_0 + (1-t)\boldsymbol\Sigma_0\,\mathbf S_k^{-1}\,\boldsymbol\delta_k.
\end{align*}
Subtracting and marginalizing over $K$ yields the mixture drift in Equation~\eqref{eq:mixture_gaussian}.

\section{Additional Diagnostics}
\label{app:diagnostics}
We include auxiliary experiments that compare finite-difference and automatic-differentiation Jacobian-vector products and study sensitivity to the perturbation scale $\varepsilon$.

\begin{figure}[H]
    \centering
    \begin{subfigure}[b]{0.49\textwidth}
        \centering
        \includegraphics[width=\textwidth]{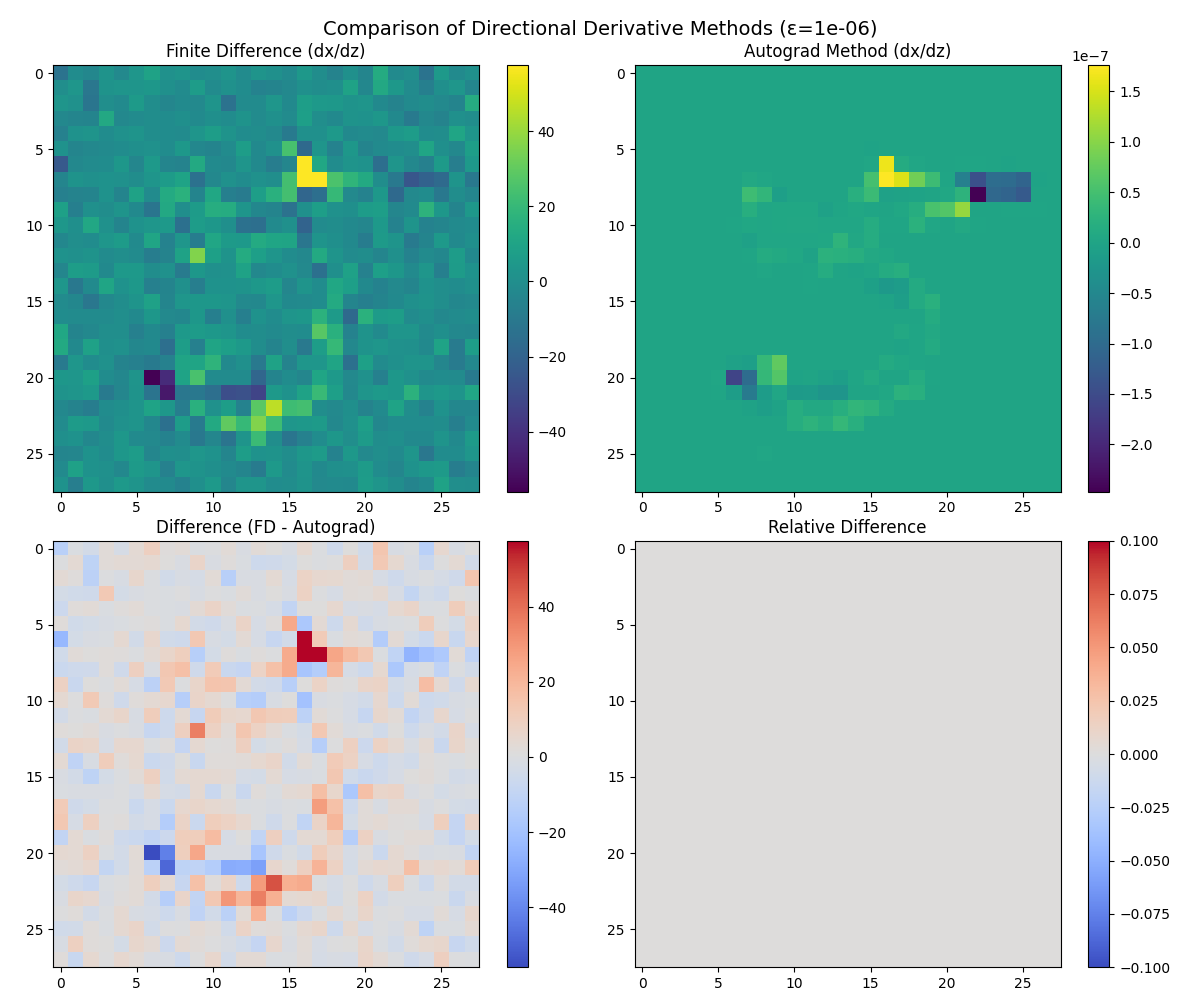}
        \caption{Finite differences vs. JVP via automatic differentiation (example).}
        \label{fig:derivative_method_comparison}
    \end{subfigure}
    \hfill
    \begin{subfigure}[b]{0.49\textwidth}
        \centering
        \includegraphics[width=\textwidth]{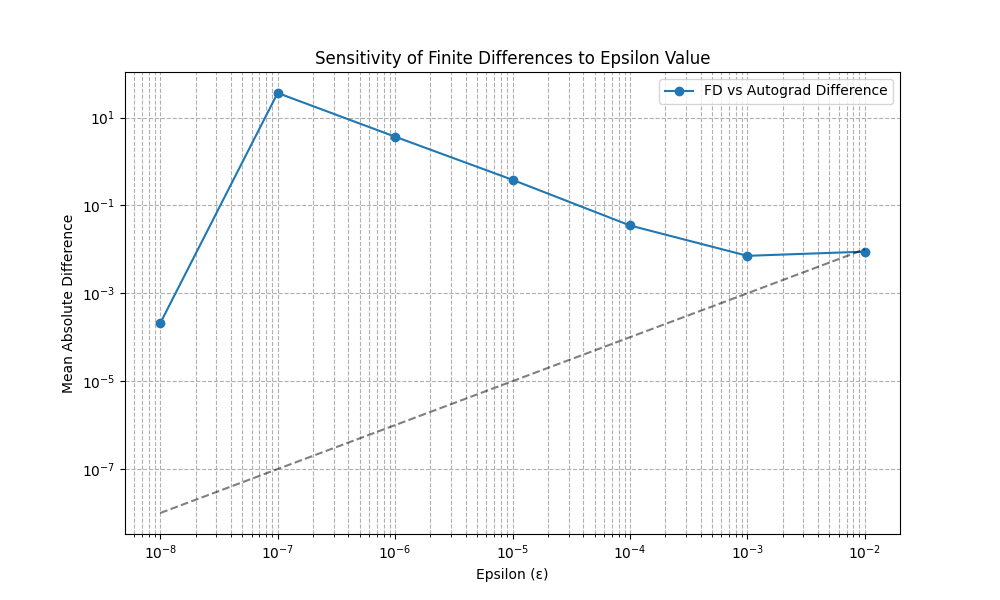}
        \caption{Sensitivity of finite differences to $\varepsilon$.}
        \label{fig:epsilon_sensitivity}
    \end{subfigure}
    \caption{Additional diagnostics for Jacobian-vector estimation.}
    \label{fig:diagnostics}
\end{figure}

\label{appendix:}

\bibliographystyle{apalike}
\bibliography{report/references}

\end{document}